\pdfoutput=1

\documentclass[11pt]{article}

\usepackage[]{EMNLP2023}
\usepackage{times}
\usepackage{latexsym}
\usepackage{multicol}
\usepackage{multirow}
\usepackage{tablefootnote}
\usepackage{graphicx}
\usepackage[T1]{fontenc}

\usepackage[utf8]{inputenc}

\usepackage{microtype}

\usepackage{inconsolata}

\title{A Simple yet Efficient Ensemble Approach for AI-generated Text Detection}

\author{Harika Abburi\textsuperscript{1}\thanks{Corresponding author}, Kalyani Roy\textsuperscript{1}, Michael Suesserman\textsuperscript{2}, Nirmala Pudota\textsuperscript{1},\\ \textbf{Balaji Veeramani\textsuperscript{2}, Edward Bowen\textsuperscript{2}, Sanmitra Bhattacharya\textsuperscript{2}} \\
\textsuperscript{1}Deloitte \& Touche Assurance \& Enterprise Risk Services India Private Limited India \\ \textsuperscript{2}Deloitte \& Touche LLP, USA \\
\texttt\{abharika, kalyroy, msuesserman, npudota,
bveeramani, edbowen, sanmbhattacharya\}@deloitte.com}

\begin{document}
\maketitle
\begin{abstract}
Recent Large Language Models (LLMs) have demonstrated remarkable capabilities in generating text that closely resembles human writing across wide range of styles and genres. However, such capabilities are prone to potential abuse, such as fake news generation, spam email creation, and misuse in academic assignments. Hence, it is essential to build automated approaches capable of distinguishing between artificially generated text and human-authored text. In this paper, we propose a simple yet efficient solution to this problem by ensembling predictions from multiple constituent LLMs. Compared to previous state-of-the-art approaches, which are perplexity-based or uses ensembles with a number of LLMs, our condensed ensembling approach uses only two constituent LLMs to achieve comparable performance. Experiments conducted on four benchmark datasets for generative text classification show performance improvements in the range of 0.5 to 100\% compared to previous state-of-the-art approaches. We also study the influence that the training data from individual LLMs have on model performance. We found that substituting commercially-restrictive Generative Pre-trained Transformer (GPT) data with data generated from other open language models such as Falcon, Large Language Model Meta AI (LLaMA2), and Mosaic Pretrained Transformers (MPT) is a feasible alternative when developing generative text detectors. Furthermore, to demonstrate zero-shot generalization, we experimented with an English essays dataset, and results suggest that our ensembling approach can handle new data effectively. 
\end{abstract}


\section{Introduction}

   
The domain of Natural Language Generation (NLG) is witnessing a remarkable transformation with the emergence of Large Language Models (LLMs) such as Generative Pre-trained Transformer (GPT-4) \cite{openai2023gpt4},  Large Language Model Meta AI (LLaMA-2) \cite{touvron2023llama}, Pathways Language Model (PaLM) \cite{chowdhery2022palm}, Bard\footnote{https://bard.google.com/}, and Text-to-Text Transfer Transformer (T5) \cite{2020t5}. LLMs, characterized by their large parameter size, have shown state-of-the-art capabilities in generating text that closely mirrors the verbosity and style of human language. They have shown exceptional performance across a wide range of applications, such as story generation \cite{fan2018hierarchical}, Artificial Intelligence (AI)-assisted writing \cite{hutson2021robo}, medical question answering \cite{kung2023performance}, conversational response generation \cite{mousavi2023response}, radiology report generation \cite{mallio2023large}, and  code auto-completion \cite{tang2023domain}. 
Moreover, their capacity to generalize across tasks without the need for explicit training (referred to as zero-shot learning) or conditioning on only a few examples (referred to as few-shot learning) have substantially reduced the need for extensive, task-specific training efforts. These capabilities have significantly lowered the barrier of integrating LLMs into various language generation applications.

With the ability to generate coherent human-like text, the LLMs can also be misused for unethical purposes, such as fake news generation \cite{uchendu2021turingbench}, phishing or spamming \cite{weiss2019deepfake}, and fabrication of product reviews \cite{gambetti2023combat}.
It has become increasingly crucial for both humans and automated systems to be able to detect and distinguish AI-generated text, particularly when this text is employed for disseminating misinformation or propaganda \cite{weidinger2021ethical}. To address these challenges, automatic detection of AI-generated text has recently become an active area of research.

Diverse modeling strategies, ranging from simple statistical techniques to cutting-edge Transformer-based architectures, have been investigated to develop solutions capable of distinguishing AI-generated text from those written by humans. \citet{gehrmann2019gltr} proposed straightforward statistical methods for identifying model-generated text that could be integrated into a visual tool to aid in their detection process. The authors assumed that AI systems produce text from a limited set of language patterns for which they have a high level of confidence. \citet{wu2023llmdet} and \citet{yang2023dnagpt} explored entropy, n-gram frequency, and  perplexity to distinguish between human-authored and AI-generated texts. Advanced deep-learning frameworks, such as Transformer-based models have also been explored to improve the precision and reliability of AI-generated text detection techniques. DetectGPT \cite{mitchell2023detectgpt} focused on generating minor perturbations of a text passage using a generic pre-trained T5 model. It then compared the log probability of both the original text and the perturbed versions to determine if the text is authored by a human or generated by AI. \citet{liu2022coco} proposed a Coherence-based Contrastive learning (CoCo) model where the input text is represented as a coherence graph to capture its entity consistency. Robustly optimized Bidirectional Encoder Representations from Transformers (BERT) approach (RoBERTa) embeddings are extracted and concatenated with sentence level graphical representations. In order to improve the model's performance, it is trained using a combination of contrastive loss and cross-entropy loss.
Most recently, \citet{abburi2023ensemble, abburi2023generative} proposed an ensemble modeling approach for detecting AI-generated text where the probabilities from various constituent pre-trained LLMs are concatenated and passed as a feature vector to machine learning classifiers. The ensemble modeling approach resulted in improved predictions compared to what any individual classifier could achieve independently.


Although the primary purpose of AI-generated text detectors is to mitigate risks associated with harmful AI-generated content, erroneously classifying genuine, human-authored work as AI-generated can, conversely, lead to significant harm.  Recently, there has been growing apprehension regarding the accuracy and reliability of these generative AI text detectors \cite{liang2023gpt, sadasivan2023can, he2023mgtbench}. \citet{liang2023gpt}  highlighted potential bias observed with several GPT detectors. The authors showed that a majority of existing detectors incorrectly classified English writing samples from non-native English speakers as AI-generated. Surprisingly, altering language created by non-native speakers with prompts like \emph{"Enhance it to sound more like that of a native speaker"} resulted in a significant drop in misclassification. 
This highlights that a majority of detectors prioritize low perplexity as a primary criterion for identifying text as AI-generated. Since the potential bias in detectors is tied to perplexity scores, the authors propose a more robust and equitable redesign of these detectors. In addition, they propose thorough evaluation of these detectors that takes other important metrics such as bias and fairness into consideration.

In this paper, we extend the work of \citet{abburi2023ensemble,abburi2023generative} by proposing an architecture that is simpler in design, while maintaining model performance. We validate the effectiveness of our model by benchmarking it on various publicly available datasets, including the Automated Text Identification (AuTexTification) \cite{autextification} dataset. 
We also study how inclusion of data generated by various LLMs in the training corpus affects the model performance and generalizability.  
In order to examine if our approach suffers from similar drawbacks and biases as other perplexity-based approaches, we evaluate the zero-shot performance of our trained model on the aforementioned English essays dataset \cite {liang2023gpt} and report the corresponding results.

In summary, our key contributions in this paper are: 1) we propose a simpler non-perplexity based AI-text detector model that extends prior work 2) we demonstrate the robustness of our approach across multiple benchmark datasets, including the one that examines potential biases in model predictions, 3) we analyze the influence that training data from individual LLMs have on model performance and 4) we find that excluding GPT data from training sets improves the accuracy of detecting human-authored samples.

\begin{table*}[!h]
\scriptsize
\centering
\begin{tabular}{p{1.1cm}lllllll}
\hline
&\textbf{Dataset}&  &  \multicolumn{2}{c}{\textbf{$Train$}}&  & \multicolumn{2}{c}{\textbf{$Test$}}\\ \cline{1-2}\cline{4-8}
&&  & \textbf{$Human$} & \textbf{$AI$} && \textbf{$Human$} & \textbf{$AI$}\\ \cline{4-5}\cline{7-8}
\multirow{6}{*}{\begin{tabular}[c]{@{}l@{}}Benchmark\\ Datasets\end{tabular}}
& AuText && 17,046 &  16,799  && 10,642  & 11,190  \\
& AA &&   213 & 1,706 && 853 & 6,822   \\ 
& TB  &&  5,964 & 10,6240 && 1,915 & 35,442  \\
& AP && 82 & 78 && 18 & 12    \\
& GPT-OD && 250,000 & 250,000 && 5,000 & 5,000\\
& EWEssays &&--&--&&394&352\\
\hline\hline
\multirow{3}{*}{\begin{tabular}[c]{@{}l@{}}Curated\\ Datasets\end{tabular}}
&D1&&17,046 & 8,263 &&10,642& 11,190\\
&D2&&17,046 &  16,799 & &10,642& 11,190\\
&D3&&17,046 &  16,799&&10,642& 11,190\\
\hline
\end{tabular} 
\caption{Dataset statistics}
\label{tab:dataset}
\end{table*}

\section{Datasets}
In this section, we provide a brief description of various publicly available benchmark datasets for AI-generated text detection. We also describe a number of datasets that we crafted and used in our experiments (henceforth referred to as curated datasets). Table \ref{tab:dataset} shows the number of human-authored (Human) and AI-generated (AI) samples available for train and test splits of each dataset.

\subsection{Publicly available benchmark datasets}
For the first set of experiments, which demonstrate the robustness of our approach, we use multiple benchmark datasets described below. 
\subsubsection{AuTexTification (AuText):}
\label{autest}
The AuText dataset \cite{autextification} consists of human-authored and AI-generated texts from five domains, where three domains (legal, wiki, tweets) are represented in the training corpus, and two different domains (reviews, news) are represented in the testing corpus. The generated text is created using six LLMs of varying parameter sizes ranging from 2B to 175B. Three of them are BigScience Large Open-science Open-access Multilingual Language Model (BLOOM) models and the other three are GPT variants: (i). bloom-1b7 (A), (ii). bloom-3b (B), (iii). bloom-7b1 (C), (iv). babbage (D), (v). curie (E), and (vi). text-davinci-003 (F).
\subsubsection{Author Attribution (AA):}
\label{AA_dataset}
The AA dataset \cite{uchendu2020authorship} consists of nine categories: human and eight LLMs generated texts. Political news articles from CNN, New York Times, and  Washington Post represent the human-authored text. The titles of these news articles written by human journalists are used as the prompts to generate the AI-generated text from eight LLMs such as Conditional Transformer Language Model (CTRL), Cross-Lingual Language Model (XLM), eXtreme Multi-Label Multi-Task Learning with a Language Model (XLNet), GPT, GPT2, Grover, Meta's Fair, and Plug and Play Language Model (PPLM).
\subsubsection{Turing Bench (TB):}
The TB dataset \cite{uchendu2021turingbench} is created by gathering around 10k news articles written by journalists in various media channels. The title of each article is used as a prompt to generate the text from 19 LLMs, such as GPT, GPT2, GPT3,  PPLM, Transformer-XL, XLM, XLNet, and various versions of these models. After preprocessing, the dataset comprises 168,612 articles with around 8k samples in each LLM category, including human-authored.


\subsubsection{Academic Publications (AP):}
The AP dataset \cite{liyanage2022benchmark} is composed of 100 papers selected from ArXiv in computation and language domain and labeled as human-authored. GPT-2 is used to generate the 100 equivalent research papers and labeled as AI-generated. From both human-authored and GPT2-generated text, the sections such as methodology, results, and discussion which contain diagrams, tables, equations are ignored.

\subsubsection{Gpt-2-Output-Dataset (GPT-OD):}
The GPT-OD \cite{radford2019language} dataset contains data from WebText test set as well as samples generated by four GPT-2 variants  (with parameters 117M, 345M, 762M, and 1542M) trained on the WebText training set. More details about the dataset can be obtained here \footnote{https://github.com/openai/gpt-2-output-dataset}. In this study, we consider 255k samples from the WebText test set as human-authored and 255k samples generated using the GPT-2 XL-1542M model (temperature 1, no truncation) as AI-generated samples.



\subsubsection{Essays from native and non-native English writers:}
This dataset is primarily comprised of essays authored by native and non-native English speakers \citet{liang2023gpt}. US 8-th grade student essays represent essays authored by native English speakers, while Test of English as a Foreign Language (TOEFL) essays obtained from a Chinese educational forum represent essays authored by non-native English speakers. ChatGPT 3.5 with simple prompts was used on the aforementioned essays, as well as Stanford CS224n final project abstracts and US Common App college admission essays, to generate artificial essays. We refer to this dataset as \emph{EWEssays} hereafter. 
In this paper, we evaluate the performance of our model on this dataset using a zero-shot approach, utilizing the complete dataset for inference.



\begin{table}[h] 
\centering
\begin{tabular}{lllrr}
    \hline 
    \textbf{Model} & & \textbf{$Train$} & \textbf{$Test$} \\ \cline{1-1}\cline{3-4}
    human-authored && 17046&10642\\
    bloom-1b7 (A) && 2,750 & 1,704\\
    bloom-3b (B)  && 2,705 & 1,782\\
    bloom-7b1 (C) && 2,808 & 1,831\\
    babbage (D)   && 2,834 & 1,960\\
    curie (E)     && 2,843 & 1,958\\
    text-davinci-003 (F)&&
                      2,859 & 1,955\\
    \hline
\end{tabular}
\caption{AuText dataset statistics}
\label{tab:aut_dataset}
\end{table}

\subsection{Curated training datasets} \label{sec:curated_dataset}
We created a number of curated datasets motivated by the following factors: 1) demonstrate the influence of training data from individual LLMs on model performance, 2) explore whether model performance is affected in out-of-domain testing, i.e., the model is tested on a dataset generated by a LLM that is not used in training data creation, and 3) 
specifically, analyze whether a model trained without GPT data can achieve similar performance to a model trained using GPT data, which is subject to specific restrictions regarding commercial usage.
While we focus primarily on the AuText dataset to derive these curated datasets, our analysis is broadly applicable to other datasets mentioned in Section 2.1.

The distribution of train and test splits for both human-authored and AI-generated data in the AuText dataset are shown in Table \ref{tab:aut_dataset}. Around half of the AI-generated data is produced by BLOOM-based models \cite{scao2022bloom}, while the rest are generated by GPT-based models.
Given the restrictions on commercial usage of data generated by GPT-based models\footnote{https://openai.com/policies/terms-of-use}, we wanted to explore whether replacing GPT data with data from other recent open LLMs (LLaMA2\footnote{https://huggingface.co/meta-llama/Llama-2-13b-chat}, Falcon\footnote{https://huggingface.co/tiiuae/falcon-40b}, and MPT \footnote{https://huggingface.co/mosaicml/mpt-30b-instruct}) is a feasible alternative for training generative text detectors. We selected LLaMA2, because it outperformed other open LLMs on various external benchmarks, including reading comprehension, reasoning, coding, and knowledge tests \cite{touvron2023llama}. The LLaMA2 chat models have additionally been trained on over 1 million human annotations compared to its previous version. Prior to LLaMA2, Falcon and MPT were outperforming other open LLMs on the open LLM leaderboard\footnote{https://huggingface.co/spaces/HuggingFaceH4/
open\_llm\_leaderboard}. 

Using the three selected open LLMs, we created the following variants of the AuText dataset:  \\
1. In the first variant (D1), we removed all GPT-based data (categories D, E, and F in Table 2) from the AuText training data.\\
2. In the second variant (D2), we replaced the training data from the GPT-based models (categories D, E, and F) with that from LLaMA2-13b-chat model. The prompts we used with the LLaMA2-13b-chat model were the same ones used by the developers of the AuText dataset.\\
3. In the third variant (D3), we substituted training data from categories D, E, and F (Table 2) with data generated from Falcon-40b-chat, MPT-30b-instruct, and LLaMA2-13b-chat, respectively. As before, we used the same prompts as those used by the developers of the AuText dataset.

In the D1 dataset, the number of training samples are reduced from 33845 to 25309 as we removed GPT-based data. In the D2 and D3 datasets, the number of samples in training data is same as the AuText training samples since we just replaced the GPT samples with the same number of samples generated using open LLMs. In all these curated training datasets, text from human-authored and  BLOOM-based models (categories A-C in Table \ref{tab:aut_dataset}) remain unchanged. No changes were made to the test datasets from AuText.


\begin{table*}[h]
\centering
\resizebox{0.8\linewidth}{!}{
\begin{tabular}{llllllll}
\hline                        
\textbf{Dataset} && \textbf{Baseline model}  && 
\textbf{$Acc$}  & \textbf{$F_{macro}$} & \textbf{$Pre$} & \textbf{$Rec$}
\\ \cline{1-1} \cline{3-3} \cline{5-8}
AA&& RoBERTa-base~\cite{uchendu2020authorship}   &&
0.970 & 0.923 & 0.932 & 0.914 \\

TB && RoBERTa-large-MNLI~\cite{uchendu2021turingbench} && 
0.997 & 0.985 & 0.976 & 0.995 \\

AP && DistilBERT~\cite{liyanage2022benchmark} && 
0.250 & 0.242 & 0.242 & 0.242 \\

GPT-OD && COCO~\cite{liu2022coco}    && 
0.943 & 0.941 &  --  &  --\\ \hline
\end{tabular} }
\caption{Performance on various benchmark datasets with state-of-the-art models.}
\label{tab:expt1}
\end{table*}

\begin{table*}[!h]
\centering
\resizebox{0.95\linewidth}{!}{
\begin{tabular}{lllllllllllll}
\hline
\multirow{2}{*}{\textbf{Dataset}} & \multicolumn{1}{c}{} & \multicolumn{2}{c}{$Acc$}   & \multicolumn{1}{c}{} & \multicolumn{2}{c}{$F_{macro}$}& \multicolumn{1}{c}{} & \multicolumn{2}{c}{$Pre$}                   & \multicolumn{1}{c}{} & \multicolumn{2}{c}{$Rec$}\\ \cline{3-4} \cline{6-7} \cline{9-10} \cline{12-13} 
   & \multicolumn{1}{c}{} & \multicolumn{1}{c}{E} & \multicolumn{1}{c}{SE} & \multicolumn{1}{c}{} & \multicolumn{1}{c}{E} & \multicolumn{1}{c}{SE} & \multicolumn{1}{c}{} & \multicolumn{1}{c}{E} & \multicolumn{1}{c}{SE} & \multicolumn{1}{c}{} & \multicolumn{1}{c}{E} & \multicolumn{1}{c}{SE} \\ \cline{1-1} \cline{3-4} \cline{6-7} \cline{9-10} \cline{12-13} 
\multirow{2}{*} {AA}&&  
0.994 &  0.990  &&  0.986  & 0.975 &&
0.993 &  0.988 &&  0.979 & 0.962) \\
&& 
(+2.5\%)& (+2.1\%) && (+6.8\%) &(+5.6\%) &&
(+6.5\%)& (+6.0\%) && (+7.1\%) &(+5.2\%) \\ \hline

\multirow{2}{*} {TB} &&
0.998 & 0.998 && 0.990 & 0.989 &&  
0.997 & 0.993 && 0.983 & 0.986\\
&& 
(+0.1\%)& (+0.1\%) && (+0.5\%) &(+0.4\%) &&
(+2.2\%)& (+1.7\%) && (-1.2\%) &(-0.9\%) \\ \hline
\multirow{2}{*} {AP} &&
0.500 & 0.475 && 0.479 & 0.475 &&
0.484 & 0.475 && 0.485 & 0.475 \\
&& 
(+100.0\%)& (+90.0\%) && (+97.9\%) &(+96.3\%) &&
(+100.0\%)& (+96.3\%) && (+100.4\%) &(+96.3\%) \\ \hline
\multirow{2}{*} {GPT-OD}  &&  
0.990 & 0.983 && 0.989 & 0.983 &&
0.990& 0.983 && 0.990 & 0.983\\
&& 
(+5.0\%)& (+4.2\%) && (+5.1\%) &(+4.4\%) &&
--& -- && -- & -- \\ \hline
\end{tabular}}
\caption{Performance of \emph{Ensemble} (E) and \emph{Short Ensemble} (SE) models on four datasets. Numbers in the parenthesis indicate percentage changes compared to baselines.}
\label{tab:expt2}
\end{table*}


\section{Approach}
We used an ensemble modeling approach similar to the one proposed by \cite{ abburi2023ensemble, abburi2023generative}, where each input is passed through five pre-trained models, namely: 1. Decoding-enhanced BERT with disentangled attention (DeBERTa) large\footnote{https://huggingface.co/microsoft/deberta-large}  \cite{he2021deberta}, 2. cross-lingual language model RoBERTa (XLM-RoBERTa) with Cross-lingual Natural Language Inference (XNLI)\footnote{https://huggingface.co/vicgalle/xlm-roberta-large-xnli-anli}, 3. RoBERTa large\footnote{https://huggingface.co/roberta-large} \cite{DBLP:journals/corr/abs-1907-11692}, 4. RoBERTa base OpenAI detector\footnote{A finetuned sequence classifier based on RoBERTa-base (125 million parameters)\footnote{https://huggingface.co/roberta-base-openai-detector} and RoBERTa-large (356 million parameters)} \cite{solaiman2019release}, and 5. XLM-RoBERTa NLI\footnote{https://huggingface.co/sentence-transformers/xlm-r-100langs-bert-base-nli-stsb-mean-tokens} \cite{reimers-2019-sentence-bert}. In the training phase, these models are fine-tuned using the training data for each dataset shown in Table \ref{tab:dataset} (except EWEssays). For inference and testing, each model independently generates classification probabilities. In order to maximize the advantage of each model, each of these probabilities are concatenated to create a feature vector and passed as an input to train a voting classifier (Logistic Regression (LR), Random Forest (RF), Gaussian Naive Bayes (NB), Support Vector machines (SVM) \cite{mahabub2020robust}) to produce final predictions. Hereafter, we refer to this architecture as \emph{Ensemble}. 

In addition to experimenting with ensembling five models proposed by \cite{ abburi2023ensemble, abburi2023generative} , we also conducted experiments with various combinations of these models using the same architecture. We observed that an ensemble of only the RoBERTa base OpenAI detector and the XLM-RoBERTa NLI model, along with the voting classifier, delivers performance comparable to the \emph{Ensemble} architecture. Henceforth, we refer to this architecture as \emph{Short Ensemble}. For both architectures, the experimental setup and hyperparameter choices are similar to \citet{abburi2023generative}.


\section{Experiments}
In this section, we present an evaluation of our AI-generated text detection experiments. Results are presented for multiple models on both benchmark and curated datasets. Results from a zero-shot evaluation on the EWEssays dataset is also presented. 
 Traditional classification metrics, namely, accuracy ($Acc$), macro F1 score ($F_{macro}$), precision ($Prec$), and recall ($Rec$) are reported for each of the experiments.

\subsection{Performance of proposed architectures across various benchmark datasets}

As baselines, we use four Transformer-based architectures, which, to our knowledge, are the current state-of-the-art models on the corresponding benchmark datasets: 1. AA dataset: RoBERTa-base~\cite{uchendu2020authorship}, 2. TB dataset: RoBERTa-large-MNLI~\cite{uchendu2021turingbench},  3. AP dataset: DistilBERT~\cite{liyanage2022benchmark}, and 4. GPT-OD dataset: COCO~\cite{liu2022coco}. RoBERTa-large-MNLI is a RoBERTa-large model fine-tuned on the Multi-Genre Natural Language Inference (MNLI) corpus. The DistilBERT model uses knowledge distillation during the pre-training phase. Both RoBERTa and DistilBERT were fine-tuned for this experiment. COCO is a coherence-based contrastive learning model that detects AI-generated texts in low-resource settings.

Table~\ref{tab:expt1} shows the results of baseline models on four publicly available benchmark datasets. 
The results from the \emph{Ensemble} and \emph{Short Ensemble} models on the same four benchmark datasets are shown in Table~\ref{tab:expt2}. The results show both the \emph{Ensemble} and \emph{Short Ensemble} architectures perform well when compared to the other architectures across datasets. In both ensemble architectures the RoBERTa base OpenAI detector model shows stronger performance compared to other constituent models. When compared to the state-of-the-art, we find that our models deliver performance improvements in the range of 0.5-97.9\% for $F_{macro}$, across the benchmark datasets. The highest improvement is on the AP dataset (97.9\%), followed by the AA dataset (6.8\%). Notably, the results from the \emph{Short Ensemble} model closely approximate those from the \emph{Ensemble} architecture. This indicates that an ensemble of just two models, as seen in the \emph{Short Ensemble} architecture is adequate to achieve state-of-the-art performance. Importantly, this approach significantly simplifies the model's complexity compared to the larger ensemble model. As a result, we choose the \emph{Short Ensemble} architecture for presenting the remaining results in the paper.

\subsection{Analysis of model performance on curated training datasets}
Each of the curated training datasets (D1, D2 and D3) are variants of the AuText dataset, and comprises data from different combinations of LLMs. Table \ref{tab:expt3} illustrates the influence these different datasets have on model performance compared to the unmodified AuText training data.  The \emph{Short Ensemble} architecture is trained on each of the curated datasets independently and evaluated on the AuText test set.
Based on the results from D1, D2, and D3, it is evident that, despite the absence of text generated by GPT models in the training data, the \emph{Short Ensemble} model is able to effectively detect GPT-generated text. In D1, by simply removing the GPT text from the training data the $F_{mac}$ score improved to 0.769 from its baseline $F_{mac}$ 0.732. The model performance is further improved with a $F_{mac}$ score of 0.774 when GPT text is replaced with LLaMA2 data. 
An interesting observation we made regarding the D3 dataset (which uses Falcon, MPT, and LLaMA2 data) is that, even though the model demonstrated improved performance compared to AuText, it did not outperform the results achieved by exclusively utilizing text from LLaMA2.
Across these metrics, the \emph{Short Ensemble} model, fine-tuned on D2 dataset with LLaMA2 data, outperformed those trained on the AuText data and other variants.


\begin{table}[h]
\centering
\begin{tabular}{llllll}
\hline
\textbf{Dataset} &&  \textbf{$Acc$}  & \textbf{$F_{macro}$} & \textbf{$Pre$} & \textbf{$Rec$} \\   
\cline{1-1} \cline{3-6}
AuText && 
0.750 & 0.732 & 0.822 & 0.744 \\

D1  && 
0.775 & 0.769 & 0.796 & 0.771 \\

D2  && 
\textbf{0.784} & \textbf{0.774} & \textbf{0.828} & \textbf{0.779} \\ 

D3 && 
0.760 & 0.747 & 0.812 & 0.755  \\ \hline

\end{tabular} 
\caption{ Result on AuText test data with \emph{Short Ensemble} model trained on four different training sets.} 
\label{tab:expt3}
\end{table}

\begin{figure}[h]
\centering
\includegraphics[width=7.5cm]{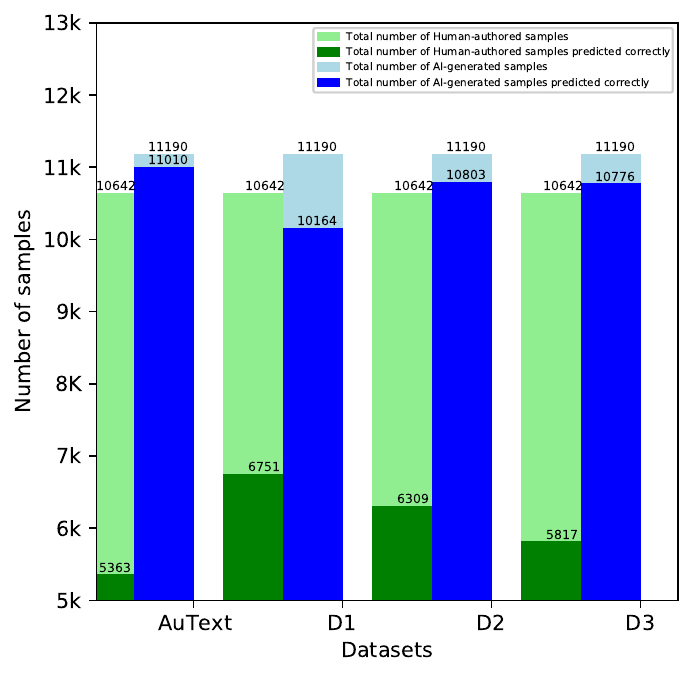}%
\caption{Performance of \emph{Short Ensemble} model on AuText and our curated datasets.}\label{fig:autext}
\end{figure}
Figure \ref{fig:autext} presents the distribution of samples within human-authored and AI-generated classes, along with the number of correct predictions across the four datasets. The results indicate that the model trained on the AuText data correctly predicts the highest number of AI-generated samples, followed by D2, D3, and D1, with only minor differences between them. In the case of human-authored class predictions, D1 showed higher performance followed by D2, D3 and AuText. Interestingly, the model not trained using GPT-generated text, i.e., D1, performed better in detecting human-authored text. We note, however, further investigation is required to understand why certain combinations of LLM training data underperform others. 

Overall, the three motivating factors behind the creation of these curated datasets (outlined in Section 2.2) were addressed with these experiments. Our experiments demonstrate that using recent open LLMs over commercially-restrictive GPT-based data is a feasible alternative in developing generative text detectors.

\begin{table}[h]
\centering
\begin{tabular}{ll llll}
\hline
\textbf{Dataset} && \textbf{$Acc$}  & \textbf{$F_{macro}$} & \textbf{$Pre$} & \textbf{$Rec$} \\ 
\cline{1-1} \cline{3-6}
AuText &&  
0.684 & 0.683 & 0.694 & \textbf{0.689} \\

D1  && 
\textbf{0.693} & \textbf{0.690} & 0.693 & \textbf{0.689} \\

D2  && 
0.655 & 0.639 & 0.720 & 0.670 \\

D3 && 
0.633 & 0.601 & \textbf{0.744} & 0.651 \\ \hline

\end{tabular}
\caption{Zero-shot results on EWEssays dataset with \emph{Short Ensemble} approach.}
\label{tab:expt4}
\end{table}

\begin{figure}[h]
\centering
\includegraphics[width=7.5cm]{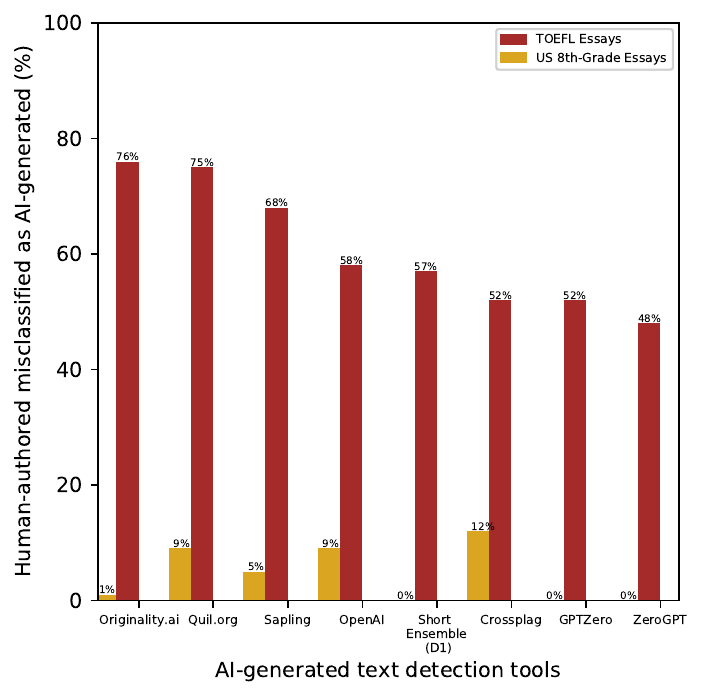}%
\caption{Performance of AI-generated text detection tools along with \emph{Short Ensemble} (D1) model on EWEssays dataset (TOEFL and US 8th grade essays categories)}\label{fig:tools}
\end{figure}

\subsection{Zero-shot generalization}
To assess the robustness and generalizability of our \emph{Short Ensemble} model, we tested the models outlined in Table \ref{tab:expt3} on the EWEssays dataset in a zero-shot setting. The results are shown in Table \ref{tab:expt4}.  We find that D1 outperforms other models achieving $F_{mac}$ score of 0.69. Furthermore, we performed an analysis to assess the precision of our model in detecting two distinct human-authored classes: US 8-th grade essays, and TOEFL essays, similar to \citet{li2023ethics}.  Figure \ref{fig:tools} depicts the performance of various generative text detectors along with \emph{Short Ensemble} (D1). Along with GPTZero\footnote{https://gptzero.me/} and ZeroGPT\footnote{https://www.zerogpt.com/}, our \emph{Short Ensemble} model with D1 dataset also classified all the US 8-th grade essays correctly as human-authored, whereas it misclassified 57.14\% of TOEFL essays as AI-generated, achieving overall accuracy of 42.86\%.  
The performance of our model in the zero-shot setting is not as promising, and highlights the need for further improvements in terms of bias and fairness evaluation. Nevertheless, we note that our model outperformed Originality.ai, Sapling.ai, Quil.org, and OpenAI text detectors as shown in Figure \ref{fig:tools}.

\section{Conclusion}
In this research, we proposed a simple yet effective \emph{Short Ensemble} model for distinguishing between AI-generated and human written text. We investigated the robustness of our proposed model across various benchmark datasets and observed that our model performs better compared to several state-of-the-art baselines. In addition, we crafted a set of datasets using open LLMs and examined their impact on model performance. Our study shows that fine-tuning models with text generated from open LLMs performs comparable or better when compared to models fine-tuned on GPT-generated text.
Furthermore, we investigated the zero-shot generalization capabilities of our model on the EWEssays dataset. We observed that our model outperformed several text detection tools in correctly classifying English essays authored by non-native English writers. However, it is important to note that our model with the highest accuracy on EWEssays achieved a score of 42.8\%, emphasizing the need for ongoing efforts to enhance both generalization and robustness in our approach. We also acknowledge that our approach should further be evaluated on detecting text generated from more advanced LLMs (LLMs with more than 175B parameters such as GPT-4 \cite{openai2023gpt4}).

\bibliography{references}
\bibliographystyle{acl_natbib}




\end{document}